\documentclass[letterpaper, 10 pt, conference]{ieeeconf}
\IEEEoverridecommandlockouts
\usepackage{stackengine}
\usepackage{amsmath}
\usepackage{amssymb}
\usepackage{xspace}
\usepackage{microtype}
\usepackage{color}
\usepackage{url}
\usepackage{lipsum}
\usepackage{fancyhdr}
\usepackage{array}
\usepackage{comment}
\usepackage{makecell}  
\usepackage{multirow}   
\usepackage{tikz}

\usepackage{adjustbox}

\usepackage{pifont}
\usepackage{cite}
\usepackage{graphicx}
\usepackage{textcomp}
\usepackage{mathtools}
\usepackage{amsmath,mleftright}
\usepackage{booktabs}
\usepackage{algpseudocode}
\usepackage{xparse}
\usepackage{comment}
\usepackage{amssymb}
\usepackage{mathbbol}
\usepackage{amsthm}
\usepackage{graphicx}
\usepackage{amsmath}
\usepackage{amssymb}

\let\labelindent\relax
\usepackage{enumitem}

\usepackage{cite}
\usepackage{tabularx}
\usepackage[table]{xcolor}
\usepackage{subfig}
\usepackage[ruled,vlined,noend]{algorithm2e}
\usepackage{float}
\usepackage{tikz}

\definecolor{mygray}{gray}{0.9}
\definecolor{codeblue}{rgb}{0,0,1}
\definecolor{rblue}{rgb}{0,0.5,1}
\definecolor{awesome}{rgb}{1.0, 0.13, 0.32}
\definecolor{hollywoodcerise}{rgb}{0.96, 0.0, 0.63}
\definecolor{lasallegreen}{rgb}{0.03, 0.47, 0.19}
\definecolor{hanpurple}{rgb}{0.32, 0.09, 0.98}
\definecolor{green(pigment)}{rgb}{0.0, 0.65, 0.31}
\definecolor{mygreen}{RGB}{93,174,86}

\newcommand{\gcircle}{%
\tikz\draw[green!70!black,fill=green!70!black] (0,0) circle (0.10) 
node[text=white,font=\bfseries\tiny] {\ding{51}};} 
\newcommand{\xcircle}{%
\tikz\draw[red,fill=red] (0,0) circle (0.10) 
node[text=white,font=\bfseries\tiny] {\ding{55}};} 
\makeatletter
\let\NAT@parse\undefined
\makeatother
\usepackage[pagebackref=false,breaklinks=true,colorlinks,bookmarks=false]{hyperref}
\hypersetup{colorlinks=true,
    linkcolor={red},
    citecolor={hanpurple},
    urlcolor={magenta}
}

\def\BibTeX{{\rm B\kern-.05em{\sc i\kern-.025em b}\kern-.08em
    T\kern-.1667em\lower.7ex\hbox{E}\kern-.125emX}}
    
\begin{document}

\title{\LARGE \bf
DepTR-MOT: Unveiling the Potential of Depth-Informed Trajectory Refinement for Multi-Object Tracking
}

\author{Buyin Deng$^{1,*}$, Lingxin Huang$^{1,*}$, Kai Luo$^{1,*}$, Fei Teng$^{1}$, and Kailun Yang$^{1,\dag}$%
\thanks{This work was supported in part by the National Natural Science Foundation of China (Grant No. 62473139), in part by the Hunan Provincial Research and Development Project (Grant No. 2025QK3019), and in part by the State Key Laboratory of Autonomous Intelligent Unmanned Systems (the opening project number ZZKF2025-2-10).}
\thanks{$^{1}$The authors are with the School of Artificial Intelligence and Robotics, Hunan University, China (email: kailun.yang@hnu.edu.cn).}%
\thanks{$^{*}$Equal contribution.}
\thanks{$^{\dag}$Corresponding author: Kailun Yang.}
}

\maketitle
\begin{abstract}
Visual Multi-Object Tracking (MOT) is a crucial component of robotic perception, yet existing Tracking-By-Detection (TBD) methods often rely on 2D cues, such as bounding boxes and motion modeling, which struggle under occlusions and close-proximity interactions. Trackers relying on these 2D cues are particularly unreliable in robotic environments, where dense targets and frequent occlusions are common. While depth information has the potential to alleviate these issues, most existing MOT datasets lack depth annotations, leading to its underexploited role in the domain. To unveil the potential of depth-informed trajectory refinement, we introduce DepTR-MOT, a category-agnostic DETR-based detector enhanced with instance-level depth information. Specifically, we propose two key innovations: (i) foundation model-based instance-level soft depth label supervision, which refines depth prediction, and (ii) the distillation of dense depth maps to maintain global depth consistency. These strategies enable DepTR-MOT to embed instance-level depth awareness into detector and to output instance-level depth during inference. All designs strictly follow the prevailing Tracking-By-Detection (TBD) paradigm, requiring no additional training or inference stages. By incorporating depth cues, our method enhances the robustness of the TBD paradigm and effectively resolves occlusion and close-proximity challenges. Experiments on the QuadTrack, DanceTrack, SportsMOT, and VisDrone19 datasets demonstrate the effectiveness of our approach, achieving HOTA scores of $27.59$, $44.939$, $61.2$, and $42.5$, respectively. These consistent gains across aerial, sports, and surveillance platforms highlight the advantages of our method in handling occlusion and close-proximity challenges in robotic tracking. The source code will be made publicly available at \url{https://github.com/warriordby/DepTR-MOT}.
\end{abstract}

\section{Introduction}

Multi-Object Tracking (MOT) is a core task in the perception of autonomous driving and mobile robots~\cite{hu2023_uniad}. Its goal is to establish reliable data associations across consecutive frames, thereby achieving accurate detection and continuous trajectory tracking in video sequences~\cite{li2025review}.
However, real-world MOT remains challenging due to frequent occlusions and close-proximity interactions between targets~\cite{milan2016mot16,sun2022dancetrack}. Since association is typically performed using 2D IoU~\cite{yang2023hard,zhao2025detrack}, such cues become unreliable in these scenarios, often resulting in trajectory fragmentation and identity switches~\cite{di2025hybridtrack}.

To address these challenges, leveraging explicit 3D spatial information has proven effective~\cite{zhou2018voxelnet}. 3D detection methods help alleviate appearance similarity and improve object discriminability under occlusion~\cite{weng2020ab3dmot}. However, acquiring and annotating large-scale 3D annotations is costly and time-consuming, severely limiting the applicability of 3D-based methods~\cite{zuo2024towards}. While zero-shot methods~\cite{brazil2023omni3d} allow 3D detection on 2D datasets, their performance remains suboptimal without domain-specific fine-tuning, leading to reduced detection accuracy~\cite{liu2023zero1to3}. In contrast, 2D data collection and annotation are more practical and cost-effective, making them a preferred choice for large-scale tracking tasks~\cite{sun2022dancetrack,10851814,cui2023sportsmot}. 
Thus, obtaining depth information using only 2D annotated datasets presents a highly promising avenue for improving MOT robustness, particularly in effectively handling occlusion and close-proximity interactions in real-world scenarios.

\begin{figure}[!t]
    \centering 
    \vspace{0.2cm}
    \includegraphics[width=0.48\textwidth]{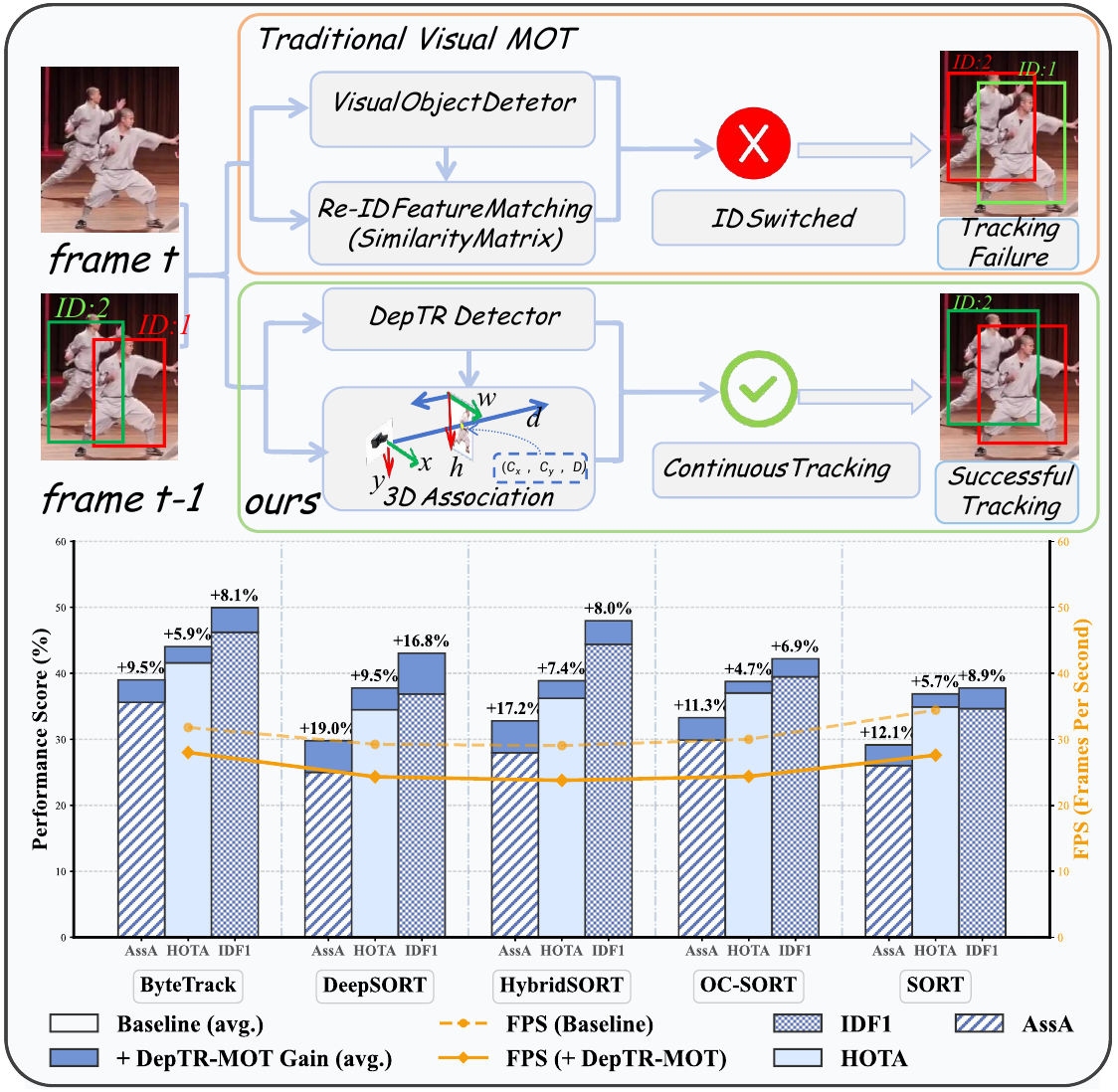} 
    \vskip-1ex

    \caption{Overview of the proposed DepTR: by cultivating intrinsic depth perception via foundation-model guided supervision, DepTR-MOT jointly regresses instance-specific absolute depth and bounding box coordinates during the detection stage, seamlessly incorporating depth cues into data association to enhance trajectory continuity.}
    \label{fig:abstract}
    \vskip-4ex
\end{figure}

Recent studies have incorporated depth cues into MOT through pseudo-depth motion modeling or geometric reasoning~\cite{wang2025pd,limanta2024camot,quach2024depth,sun2025view}, as well as monocular depth estimation or predictable depth cues~\cite{wu2024depthmot,zhao2025detrack}. Despite their effectiveness, these methods often rely on heuristic geometric assumptions, external depth networks, or camera pose estimation during inference, which may increase system complexity and limit real-time deployment. Therefore, a lightweight framework that produces MOT-specific instance-level depth cues without external depth inference is still desirable.
These limitations call for a solution that automatically yields precise, MOT-specific depth while maintaining low computational overhead and deployment ease.

To address these limitations and unlock the potential of depth-informed trajectory refinement, we propose DepTR, which leverages a lightweight detector as its core component, specifically tailored to meet the demands of Multi-Object Tracking (MOT) tasks (Fig.~\ref{fig:abstract}). Unlike conventional detectors~\cite{zhao2024detrs} that only output 2D localization, DepTR also predicts instance-level depth, providing reliable depth cues to enhance TBD frameworks. This capability stems from two key designs: (i) a foundation-guided~\cite{depth_anything_v2} soft-label supervision strategy, where instance-level depth signals are derived from pre-trained depth and segmentation models~\cite{ravi2024sam2} to supervise DepTR; and (ii) a dense depth map distillation mechanism, which enforces global scale consistency and stabilizes convergence against sparse supervision. With these designs, DepTR endows standard 2D detectors~\cite{peng2024dfine} with explicit instance-level depth awareness, while remaining lightweight and easily integrable into existing tracking pipelines~\cite{wojke2017simple}. These designs equip standard 2D detectors with depth-aware representations while preserving compatibility with existing TBD trackers.

\begin{figure}[!t]
    \centering  
    \includegraphics[width=0.48\textwidth]{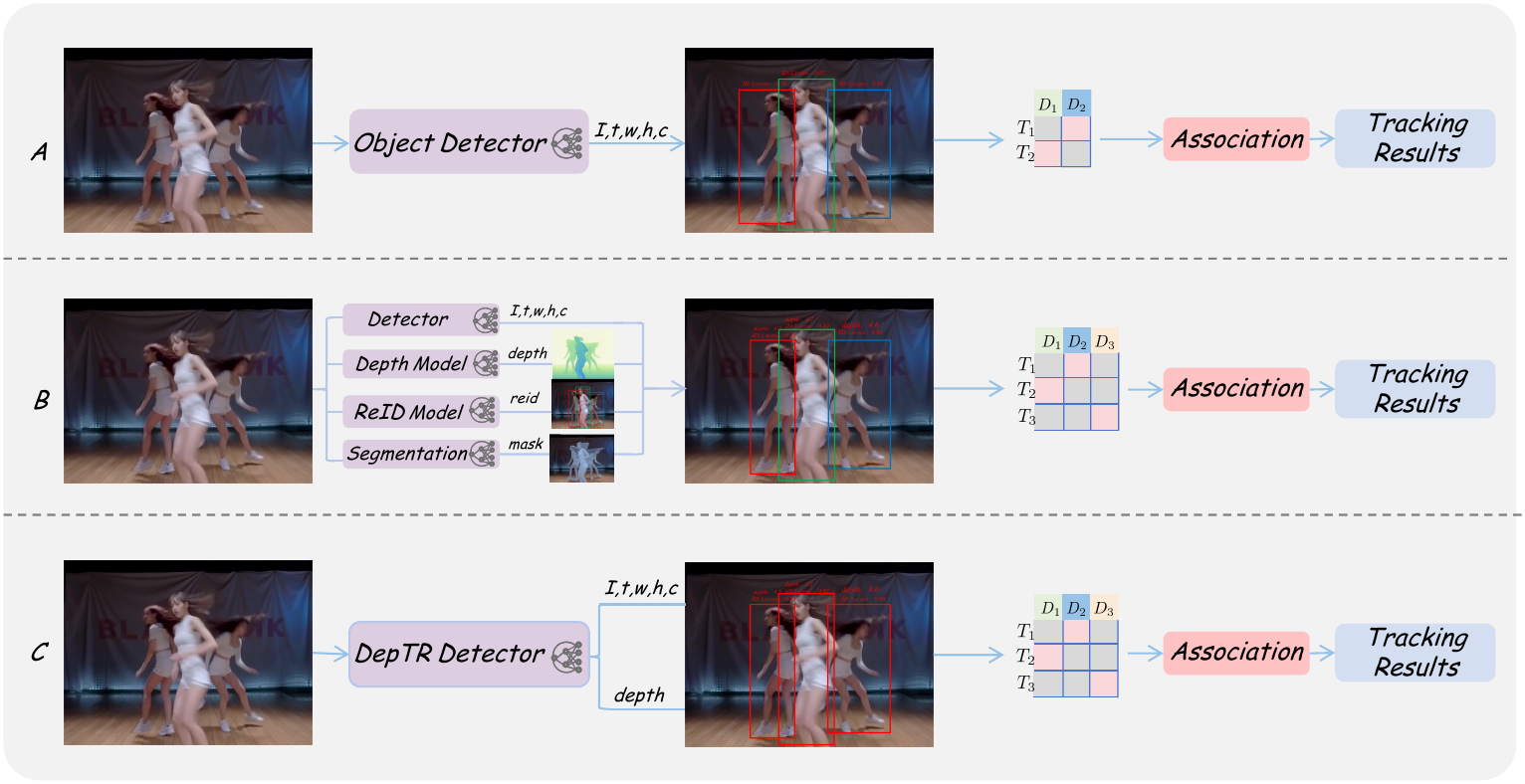} 
    \vskip-1ex
    \caption{Comparison of tracking paradigms: (A) conventional model outputs only bounding boxes and confidence scores; (B) current depth-informed MOT methods rely on external modules for additional depth or appearance features, leading to bulky inference; (C) our proposed model directly outputs instance-level depth in the detection head, achieving lightweight and depth-aware tracking with improved robustness under occlusion.}
    \label{fig:Comparison.pdf}
    \vskip-3ex
\end{figure}

We aim to establish a plug-and-play paradigm for multi-object tracking. To this end, we adopt the Tracking-By-Detection (TBD) paradigm, eschewing End-To-End (E2E) tracking frameworks in favor of TBD’s superior modularity and practical deployment advantages.
DepTR-MOT explicitly introduces depth cues in a manner that naturally aligns with the two-stage TBD framework.
Specifically, our method predicts instance-level explicit depth alongside bounding box detection, and these explicit depth cues are seamlessly integrated into the Hungarian matching process for trajectory association, without any auxiliary training stages or additional inference tasks. As a result, the proposed framework incurs negligible computational overhead while remaining fully compatible with real-time robotic deployment. As illustrated in Fig.~\ref{fig:Comparison.pdf}, the DepTR-MOT paradigm effectively unlocks the potential of depth cues for improving tracking robustness and trajectory stability, overcoming the challenges of conventional depth estimation and deployment.

We introduce DepTR, a depth-distilled MOT framework that embeds instance-level depth awareness into a DETR-style detector in a category-agnostic manner. Unlike prior tracking-by-detection methods, our detector is trained with soft depth labels from foundation models, enabling it to output target depth cues during inference without extra modules. We verify DepTR-MOT on the QuadTrack~\cite{luo2025omnidirectional}, DanceTrack~\cite{sun2022dancetrack}, SportsMOT~\cite{cui2023sportsmot}, and VisDrone19~\cite{zhu2021detection} datasets. Compared with baseline TBD trackers, our method achieves consistent improvements: an average gain of $3.155$ in IDF1, $1.516$ in MOTA, $3.167$ in AssA, $0.414$ in DetA, and $2.141$ in HOTA, with a marginal decrease of only $5.19$ FPS, demonstrating its robustness under occlusion and the significance of depth cues in advancing MOT, both in robotic scenarios and in general tracking benchmarks.

The main contributions of this work are summarized as:
\begin{itemize}
\item[\textbf{·}] We introduce DepTR-MOT, a joint depth-aware tracking framework. DepTR is a plug-and-play extension for the TBD paradigm. As a lightweight and non-invasive framework, it seamlessly integrates into existing pipelines while preserving the standard detection-and-association formulation.

\item[\textbf{·}] We design a Promptable Depth Estimation Model (PEDM) and a pretrained-model-guided knowledge distillation scheme, which together form a depth-label-free training strategy, endowing models with the capability to learn and internalize intrinsic instance-level depth perception solely from 2D annotations.

\item[\textbf{·}] Evaluations across quadruped robots, aerial drones, handheld and static camera systems demonstrate DepTR's robustness and target discriminability. It mitigates occlusion and close-proximity interference, ensuring stable tracking in complex scenarios.
\end{itemize}

\begin{figure*}[!t]
\centering 
\centerline{\includegraphics[width=0.99\linewidth]{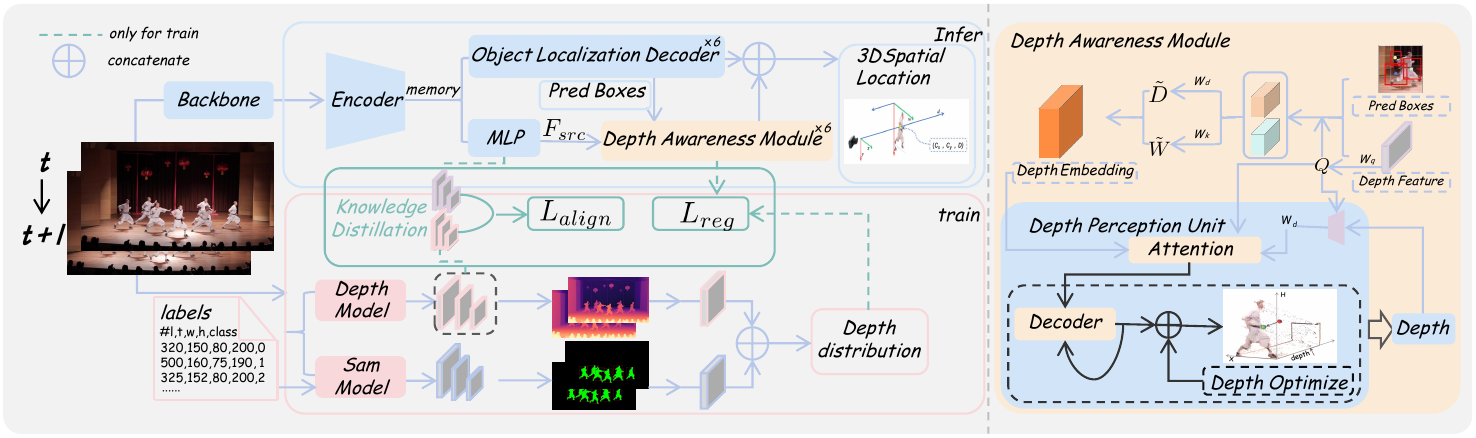}}
\vskip-1ex
\caption{Pipeline of DepTR-MOT. A prompt-based knowledge-guided foundation module leverages frozen SAM and Depth models to generate instance-level depth labels from 2D box annotations, providing label supervision and depth feature distillation to align the decoding-depth feature map during training. During inference, DepTR directly predicts 3D spatial locations using stacked object and depth-awareness blocks.}
\label{fig:pipeline}
\vskip-4ex
\end{figure*}

\section{Related Work}

Multi-Object Tracking (MOT) systems mainly consist of two key modules: Detection and Tracking. 
These two modules play distinct yet complementary roles, together forming the complete MOT pipeline.

\subsection{Detection Methods}
Modern object detection has evolved through three paradigms: two-stage, one-stage, and end-to-end Transformer detection.
Two-stage methods, from Girshick~\textit{et al.}~\cite{girshick2014rich} and Ren~\textit{et al.}~\cite{ren2016faster}, were refined by DSFSN~\cite{liu2025dual} and SSD~\cite{liu2016ssd} for efficiency and robustness.
One-stage methods focus on speed, with Redmon~\textit{et al.}~\cite{redmon2016you} proposing single-shot detection and GridCLIP~\cite{lin2025gridclip} enhancing it via super-resolution and vision–language models.
End-to-end Transformer detection began with DETR~\cite{carion2020end}, whose set-based Hungarian matching inspired Deformable DETR~\cite{zhu2020deformable}, D-FINE~\cite{peng2024dfine}, ViDT~\cite{song2021vidt}, and RT-DETR~\cite{zhao2024detrs} to improve convergence, accuracy–efficiency trade-offs, and multi-scale real-time modeling.
However, most detectors remain 2D-only, lacking depth modeling and thus prone to errors under occlusion or close proximity.
We address this by adding instance-level depth perception, providing depth cues that strengthen association and enhance tracking robustness.

\subsection{Tracking Methods}
Current MOT methods follow the Tracking-by-Detection (TBD) paradigm, evolving from motion-based matching to multi-cue fusion. Earlier approaches like SORT~\cite{wojke2017simple} and OC-SORT~\cite{cao2023observation} rely on motion priors via the Kalman Filter~\cite{welch1995introduction} and perform assignment using the Hungarian algorithm~\cite{kuhn1955hungarian} based on IoU matching~\cite{yang2023hard}, while DeepSORT~\cite{wojke2017simple} and DiffMOT~\cite{lv2024diffmot} incorporate ReID appearance features. Subsequent methods such as ByteTrack~\cite{zhang2022bytetrack} and HybridSORT~\cite{yang2024hybrid} further exploit confidence levels and geometric cues to refine association. Recently, depth information has been introduced as a critical 3D geometric cue to resolve spatial overlaps in 2D projections~\cite{wang2025pd,limanta2024camot,quach2024depth}.  However, despite their reliance on sophisticated auxiliary networks, current models lack intrinsic instance-level depth perception. Most of them necessitate computationally intensive pseudo-depth estimation or auxiliary depth modules during inference, which introduces significant computational redundancy. This forced integration fundamentally disrupts the decoupled efficiency of the Tracking-by-Detection (TBD) paradigm, leading to poor architectural scalability. Consequently, the resulting high overhead and structural complexity render these approaches ill-suited for real-time deployment on resource-constrained edge devices.

\subsection{Depth-Informed Trajectory Refinement}
Existing methods leverage depth information to improve tracking association by estimating the overall depth of regions through extra depth-map-predicted modules. DepthMOT~\cite{wu2024depthmot} and ViewTrack~\cite{sun2025view} exploit monocular depth estimation and view-adaptive strategies to enhance association accuracy under occlusion; Khanchi~\textit{et al.}~\cite{khanchi2025depth} combine depth scoring with hierarchical alignment; and DETrack~\cite{zhao2025detrack} treats depth as an additional dimension in data association. However, these approaches suffer from a significant modality gap between 2D regional representations and 3D spatial reasoning. By merely extracting depth from 2D bounding boxes via external modules, they cannot effectively filter out interference from non-target backgrounds or overlapping objects, leading to corrupted identity verification cues (as shown in Fig.~\ref{fig:sam.pdf}). 

To overcome these challenges, we propose DepTR-MOT, a computationally efficient framework that seamlessly incorporates depth into the two-stage multi-object tracking pipeline. Unlike prior works, DepTR-MOT regresses instance-specific depth cues concurrently with bounding box coordinates during the detection stage. granting the model intrinsic depth perception. In the subsequent tracking stage, this depth information is integrated into a depth-aware Hungarian matching process—where depth-based costs are explicitly added to resolve identity ambiguities. DepTR-MOT maintains the elegant scalability of two-stage trackers while significantly boosting performance.

\section{Methodology}
We propose DepTR-MOT, a distillation-based paradigm that equips 2D detectors with object-level depth perception by distilling depth supervision from pretrained foundation models, thereby enhancing MOT. During training, soft depth labels are generated from foundation models~\cite{video_depth_anything,ravi2024sam2} to supervise instance-level depth learning. Specifically, we introduce two complementary losses: a cosine-similarity loss between predicted and teacher depth features to enforce distributional consistency, and an MSE loss between predicted and reference depth values to improve depth accuracy. During inference, DepTR operates independently of teacher models and performs target-specific depth estimation by using detection boxes and their centers as spatial priors for deformable attention and depth decoding.

In Sec.~\ref{subsec:promptable_depth_estimation}, we describe the Promptable Depth Estimation Model (PEDM), which obtains stable 3D guidance from 2D annotation prompts for pretrained-model-guided distillation. Sec.~\ref{subsec:Mechanisms_of_Depth_Decoding} introduces the target-aware depth decoding strategy, where anchor-based position prediction and depth-informed sampling refine instance-level depth estimation. Finally, Sec.~\ref{subsec:Depth_Aware_Association} presents a lightweight depth-aware association strategy for applying DepTR to downstream MOT.

\subsection{Promptable Depth Estimation Model}
\label{subsec:promptable_depth_estimation}

To obtain high-quality depth supervision signals, we design PEDM to extract the 3D depth information corresponding to each target. 
This approach effectively mitigates background interference and occlusion while producing fine-grained, instance-level depth signals and providing dense feature distribution guidance. 
The entire process is divided into three stages: (1) depth estimation for consecutive video frames, (2) prompt-based target mask alignment, and (3) depth feature distillation guidance. Through this pipeline, the depth of the target is derived directly from the original 2D bounding box.

\begin{table}[!t]
\centering
\scalebox{0.9}{%
\begin{tabular}{l|c|c|c|c|c|c}
\Xhline{1.05pt}
\rowcolor{mygray}
\textbf{Setting} & Depth Cue  & \textbf{HOTA$\uparrow$} & \textbf{IDF1$\uparrow$} & \textbf{MOTA$\uparrow$} & \textbf{DetA$\uparrow$} & \textbf{AssA$\uparrow$} \\
\Xhline{1.05pt}

\multirow{2}{*}{\shortstack{$L$=1 \\ $\rho$=1}} & \xcircle & 23.095  & 22.043 & -5.3204 & 29.049 & 18.977  \\
                                                   & \gcircle & 26.718  & 26.525 & -3.9969 & 29.576 & 25.120  \\
\midrule

\multirow{2}{*}{\shortstack{$L$=2 \\ $\rho$=1}} & \xcircle & 23.666  & 22.201 & -2.6575 & 30.044 & 19.342 \\
                                                   & \gcircle & 27.590  & 28.035 & -3.6548 & 29.957 & \textbf{26.553} \\
\midrule                   

\multirow{2}{*}{\shortstack{$L$=4 \\ $\rho$=2}} & \xcircle & 25.402  & 25.711 & \textbf{5.3257} & 30.953 & 21.651  \\
                                                   & \gcircle & \textbf{27.673} & \textbf{28.556} & 3.2172 & \textbf{30.994} & 25.882  \\
\midrule

\multirow{2}{*}{\shortstack{$L$=6 \\ $\rho$=3}} & \xcircle & 23.376  & 23.141 & -1.0901 & 29.174 & 19.429 \\
                                                   & \gcircle & 26.364  & 27.244 & -1.1352 & 29.684 & 24.388 \\
\midrule

\multirow{2}{*}{\shortstack{$L$=8 \\ $\rho$=4}} & \xcircle & 24.317  & 24.789 & 0.2069  & 29.366 & 20.881 \\
                                                   & \gcircle & 26.944  & 28.510 & 0.0557  & 29.698 & 25.264 \\
\Xhline{1.05pt}
\end{tabular}}
\caption{Comparison of different data loading strategies on the DanceTrack dataset. Here, $L$ denotes the length of the data loading window, while $\rho$ represents the sliding interval between two consecutive windows.}
\label{tab:ablation_window_length}
\vspace{-4mm}
\end{table}

\noindent\textbf{(1) Depth estimation for consecutive video frames.} 
Temporally consistent depth estimation across video sequences is essential for providing reliable cues for trajectory association in multi-object tracking. To address this, we adopt Video Depth Anything~\cite{video_depth_anything}, a powerful model capable of generating consistent depth maps efficiently over arbitrary-length videos. Furthermore, we employ a sliding-window strategy that samples consecutive frames within each batch, reinforcing temporal continuity and providing temporal context through overlapping sequences. Specifically, with a window size $L$ and a stride $\rho$, each batch samples $T=L$ consecutive frames, where the interval between adjacent sequences is $S$ frames. As shown in Tab.~\ref{tab:ablation_window_length}, our ablation study validates the inherent stability brought by this loader strategy.

Formally, given an image sequence $\mathbf{X} \in \mathbb{R}^{B \times T \times 3 \times H \times W}$, the depth estimation network $f_{\phi}$ processes each sequence and outputs the corresponding depth maps, as shown in Eq.~\ref{equation}.
\begin{equation}
D = f_{\phi}({X})
= \{\mathbf{D}_{b,t} \mid 1 \le b \le B, 1 \le t \le T\} . \label{equation}
\end{equation}

By flattening the input $\mathbf{X}$ along the batch and temporal axes, we obtain a sequence representation of length $B \times T$, where $\mathbf{D}_{b,t} \in \mathbb{R}^{1 \times H \times W}$ denotes the predicted depth map for each corresponding frame. In this way, the network transforms RGB video inputs into temporally coherent depth sequences, thereby providing stable 3D supervision for downstream tasks.

\begin{figure}[!t]
    \centering 
    \hspace*{-0.2cm}
    \includegraphics[width=0.49\textwidth]
    {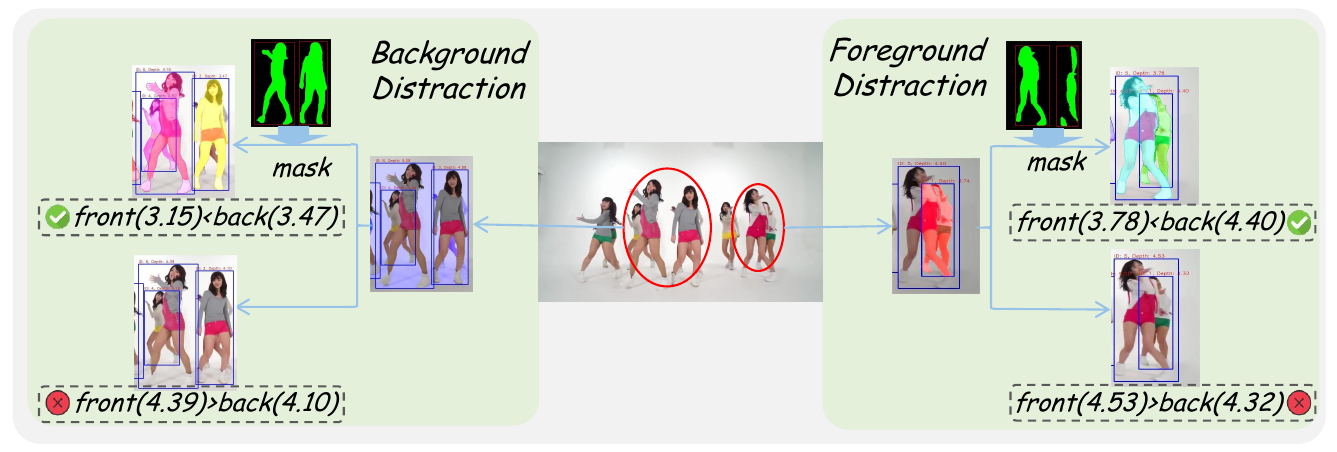} %
    \vskip-1ex
    \caption{Illustration of the pattern that bounding-box-based depth indexing is susceptible to background interference (left) and neighboring individuals (right), leading to biased or inconsistent depth values. Incorporating SAM2 mitigates these issues by applying instance masks, which remove environmental depth values and yield more accurate depth distributions.}
    \label{fig:sam.pdf}
    \vskip-4ex
\end{figure}

\noindent\textbf{(2) Prompt-based target mask alignment.} 
The existing method~\cite{khanchi2025depth} divides depth maps into regions based on bounding boxes and treats these regions as depth features of different targets for motion detection. 
However, in complex scenarios with occlusion and overlap, the depth maps within different targets’ bounding boxes may become nearly indistinguishable. 
Moreover, as shown in Fig.~\ref{fig:sam.pdf}, due to environmental scale effects, depth interference signals from surrounding objects are often much stronger than the true depth information of the targets themselves, leading to significant fluctuations in the depth maps of entire bounding boxes. 
Therefore, simply segmenting depth maps based on bounding boxes fails to provide effective supervisory signals or to support an efficient framework. 
To address this issue, we introduce instance-level segmentation information~\cite{ravi2024sam2}, which provides pixel-wise masks to eliminate both background interference and foreground occlusion, thereby yielding stable and reliable instance-level depth information.

\begin{table*}[t!]
\centering 
\noindent 

\setlength{\tabcolsep}{2.3pt} 
\renewcommand{\arraystretch}{0.8}
\setlength{\extrarowheight}{-1.2pt}
\begin{tabular}{c|c|c|c|c|c|c|c|c|c}
\cline{1-10}
\Xhline{1.05pt}
\rowcolor{mygray}
\noalign{\setlength{\extrarowheight}{-1pt}}\rule{0pt}{2.4ex}
Tracker & Dataset & w/o Depth & Detector & HOTA $\uparrow$ & IDF1 $\uparrow$ & MOTA $\uparrow$ & DetA $\uparrow$ & AssA $\uparrow$ & FPS $\uparrow$ \\
\noalign{\setlength{\extrarowheight}{-1pt}}
\Xhline{1.05pt}
\cline{1-10}
\multirow{8}{*}[-3ex]{HybridSORT~\cite{di2025hybridtrack}}

& \multirow{2}{*}{QuadTrack}
& \xcircle & DFINE & 16.663 & 15.065 & -10.439 & 21.314 & 13.557 & 33.773 \\   [ -0.6ex]
& & \gcircle & DepTR & 18.882 \textcolor{green}{(+2.22)} & 18.03\textcolor{green}{(+2.97)} & -5.750\textcolor{green}{(+4.69)} & 18.92\textcolor{gray}{(-2.39)} & 19.525\textcolor{green}{(+5.97)} & 26.232\textcolor{gray}{(-7.54)} \\
\cline{2-10}

& \multirow{2}{*}{SportsMOT}
& \xcircle & DFINE & 47.785 & 49.403 & 82.306 & 69.598 & 32.973 & 34.389 \\   [ -0.6ex]
& & \gcircle & DepTR 
& 52.919 \textcolor{green}{(+5.13)} 
& 57.119 \textcolor{green}{(+7.72)} 
& 83.043 \textcolor{green}{(+0.74)} 
& 69.984 \textcolor{green}{(+0.39)} 
& 40.203 \textcolor{green}{(+7.23)} 
& 27.393 \textcolor{gray}{(-6.99)} \\
\cline{2-10}

& \multirow{2}{*}{VisDrone}
& \xcircle & DFINE & 35.490 & 39.776 & 30.028 & 35.661 & 37.696 & 23.392 \\   [ -0.6ex]
& & \gcircle & DepTR 
& 37.145 \textcolor{green}{(+1.66)} 
& 43.087 \textcolor{green}{(+3.31)} 
& 29.787 \textcolor{gray}{(-0.24)} 
& 34.911 \textcolor{gray}{(-0.75)} 
& 41.791 \textcolor{green}{(+4.10)} 
& 20.876 \textcolor{gray}{(-2.52)} \\
\cline{2-10}

& \multirow{2}{*}{DanceTrack}
& \xcircle & DFINE & 44.895 & 73.341 & 86.983 & 44.909 & 27.644  & 24.63 \\   [ -0.6ex]
& & \gcircle & DepTR 
& 46.563 \textcolor{green}{(+1.66)} 
& 73.613\textcolor{green}{(+0.18)} 
& 88.012 \textcolor{green}{(+1.93)} 
& 46.826 \textcolor{green}{(+1.92)} 
& 29.627  \textcolor{green}{(+1.98)} 
& 20.6 \textcolor{gray}{(-3.03)} \\
\Xhline{1.05pt}
\cline{1-10}
\multirow{8}{*}[-3ex]{SORT~\cite{Bewley2016_sort}}

& \multirow{2}{*}{QuadTrack}
& \xcircle & DFINE & 15.845 & 14.082 & -3.668 & 22.777 & 11.394 & 39.459 \\   [ -0.6ex]
& & \gcircle & DepTR & 16.126\textcolor{green}{(+0.28)} & 14.729\textcolor{green}{(+0.65)} & -3.400\textcolor{green}{(+0.27)} & 22.44\textcolor{gray}{(-0.34)} & 11.933\textcolor{green}{(+0.54)} & 30.992\textcolor{gray}{(-8.46)} \\
\cline{2-10}

& \multirow{2}{*}{SportsMOT}
& \xcircle & DFINE & 47.785 & 49.403 & 82.306 & 69.598 & 32.973 & 39.500 \\   [ -0.6ex]
& & \gcircle & DepTR 
& 52.919 \textcolor{green}{(+5.13)} 
& 57.119 \textcolor{green}{(+7.72)} 
& 83.043 \textcolor{green}{(+0.74)} 
& 69.984 \textcolor{green}{(+0.39)} 
& 40.203 \textcolor{green}{(+7.23)} 
& 30.207 \textcolor{gray}{(-9.29)} \\
\cline{2-10}

& \multirow{2}{*}{VisDrone}
& \xcircle & DFINE & 35.490 & 39.776 & 30.028 & 35.661 & 37.696 & 32.998 \\   [ -0.6ex]
& & \gcircle & DepTR 
& 37.144 \textcolor{green}{(+1.65)} 
& 43.080 \textcolor{green}{(+3.30)} 
& 29.785 \textcolor{gray}{(-0.24)} 
& 34.909 \textcolor{gray}{(-0.75)} 
& 41.793 \textcolor{green}{(+4.10)} 
& 27.152 \textcolor{gray}{(-5.85)} \\
\cline{2-10}

& \multirow{2}{*}{DanceTrack}
& \xcircle & DFINE & 40.484 & 35.356 & 86.980 & 75.154 & 22.015 & 25.75
 \\   [ -0.6ex]
& & \gcircle & DepTR 
& 41.331 \textcolor{green}{(+0.85)}
& 36.089 \textcolor{green}{(+0.73)}
& 88.135 \textcolor{green}{(+1.16)}
& 76.006 \textcolor{green}{(+0.85)}
& 22.693 \textcolor{green}{(+0.68)}
& 22.05 \textcolor{gray}{(-3.7)}\\
\Xhline{1.05pt}
\cline{1-10}
\multirow{8}{*}[-3ex]{OC-SORT\cite{cao2023observation}}

& \multirow{2}{*}{QuadTrack}
& \xcircle & DFINE & 16.235 & 14.973 & -11.482 & 20.733 & 13.169 & 34.573 \\   [ -0.6ex]
& & \gcircle & DepTR & 18.872\textcolor{green}{(+2.64)} & 18.015\textcolor{green}{(+3.04)} & -5.760\textcolor{green}{(+5.72)} & 18.901\textcolor{gray}{(-1.83)} & 19.523\textcolor{green}{(+6.35)} & 27.038\textcolor{gray}{(-7.53)} \\
\cline{2-10}

& \multirow{2}{*}{SportsMOT}
& \xcircle & DFINE & 49.874 & 54.799 & 81.078 & 68.142 & 36.653 & 35.524 \\   [ -0.6ex]
& & \gcircle & DepTR 
& 54.133 \textcolor{green}{(+4.26)} 
& 61.558 \textcolor{green}{(+6.76)} 
& 81.738 \textcolor{green}{(+0.66)} 
& 68.373 \textcolor{green}{(+0.23)} 
& 43.006 \textcolor{green}{(+6.35)} 
& 27.982 \textcolor{gray}{(-7.54)} \\
\cline{2-10}

& \multirow{2}{*}{VisDrone}
& \xcircle & DFINE & 37.957 & 45.416 & 29.617 & 35.288 & 42.961 & 24.542 \\   [ -0.6ex]
& & \gcircle & DepTR 
& 38.555 \textcolor{green}{(+0.60)} 
& 46.904 \textcolor{green}{(+1.49)} 
& 29.489 \textcolor{gray}{(-0.13)} 
& 34.651 \textcolor{gray}{(-0.64)} 
& 44.677 \textcolor{green}{(+1.72)} 
& 21.036 \textcolor{gray}{(-3.51)} \\
\cline{2-10}

& \multirow{2}{*}{DanceTrack}
& \xcircle & DFINE & 44.050 & 42.712 & 85.643 & 72.953 & 26.773 & 25.35
 \\   [ -0.6ex]
& & \gcircle & DepTR 
& 43.483 \textcolor{gray}{(-0.57)}
& 42.352 \textcolor{gray}{(-0.36)}
& 86.928 \textcolor{green}{(+1.29)}
& 73.525 \textcolor{green}{(+0.57)}
& 25.877 \textcolor{gray}{(-0.90)}
& 21.43 \textcolor{gray}{(-3.9)}\\
\Xhline{1.05pt}
\cline{1-10}
\multirow{8}{*}[-3ex]{ByteTrack~\cite{zhang2022bytetrack}}

& \multirow{2}{*}{QuadTrack}
& \xcircle & DFINE & 23.903 & 23.072 & -7.935 & 29.167 & 20.364 & 38.392 \\   [ -0.6ex]
& & \gcircle & DepTR & 27.590\textcolor{green}{(+3.69)} & 28.035\textcolor{green}{(+4.96)} & -3.654\textcolor{green}{(+4.28)} & 29.957\textcolor{green}{(+0.79)} & 26.553\textcolor{green}{(+6.19)} & 30.473\textcolor{gray}{(-7.92)} \\
\cline{2-10}

& \multirow{2}{*}{SportsMOT}
& \xcircle & DFINE & 57.7 & 64.2 & 89.0 & 72.9 & 45.8 & 33.700 \\   [ -0.6ex]
& & \gcircle & DepTR 
& 61.2 \textcolor{green}{(+3.5)} 
& 69.4 \textcolor{green}{(+5.2)} 
& 90.3 \textcolor{green}{(+1.3)} 
& 74.6 \textcolor{green}{(+1.7)} 
& 50.3 \textcolor{green}{(+4.5)} 
& 30.290 \textcolor{gray}{(-3.41)} \\
\cline{2-10}
& \multirow{2}{*}{VisDrone}
& \xcircle & DFINE & 42.4 & 51.7 & 26.4 & 37.7 & 49.6 & 28.570 \\   [ -0.6ex]
& & \gcircle & DepTR 
& 42.5 \textcolor{green}{(+0.1)} 
& 52.4 \textcolor{green}{(+0.7)} 
& 27.6 \textcolor{green}{(+1.2)} 
& 37.3 \textcolor{gray}{(-0.4)} 
& 50.0 \textcolor{green}{(+0.4)} 
& 28.610 \textcolor{green}{(+0.4)} \\
\cline{2-10}
& \multirow{2}{*}{DanceTrack}
& \xcircle & DFINE & 42.343 & 45.922 & 83.352 & 67.691 & 26.640 & 26.51 \\   [ -0.6ex]
& & \gcircle & DepTR 
& 44.939 \textcolor{green}{(+2.60)}
& 49.956 \textcolor{green}{(+4.03)}
& 86.271 \textcolor{green}{(+2.92)}
& 69.687 \textcolor{green}{(+2.00)}
& 29.133 \textcolor{green}{(+2.49)}
& 22.64 \textcolor{gray}{(-3.87)} \\
\Xhline{1.05pt}
\cline{1-10}
\multirow{8}{*}[-3ex]{DeepSORT~\cite{wojke2017simple}}

& \multirow{2}{*}{QuadTrack}
& \xcircle & DFINE &15.461 & 12.514 & -8.915 & 26.135 & 9.54 & 33.93\\   [ -0.6ex]
& & \gcircle & DepTR 
& 24.325 \textcolor{green}{(+8.86)}
& 26.619 \textcolor{green}{(+14.10)}
& -27.756 \textcolor{gray}{(-18.94)}
& 25.264 \textcolor{gray}{(-0.87)}
& 24 \textcolor{green}{(+15.46)}
& 28.23 \textcolor{gray}{(-5.70)}\\
\cline{2-10}

& \multirow{2}{*}{SportsMOT}
& \xcircle & DFINE & 49.737 & 58.607 & 81.155 & 66.010 & 37.586 & 33.738 \\   [ -0.6ex]
& & \gcircle & DepTR 
& 51.456 \textcolor{green}{(+1.72)} 
& 60.711 \textcolor{green}{(+2.10)} 
& 82.308 \textcolor{green}{(+1.15)} 
& 66.799 \textcolor{green}{(+0.79)} 
& 39.760 \textcolor{green}{(+2.17)} 
& 26.861 \textcolor{gray}{(-6.88)} \\
\cline{2-10}
& \multirow{2}{*}{VisDrone}
& \xcircle & DFINE & 38.492 & 46.629 & 23.032 & 43.612 & 35.968 & 24.236 \\   [ -0.6ex]
& & \gcircle & DepTR 
& 38.538 \textcolor{green}{(+0.05)} 
& 47.030 \textcolor{green}{(+0.40)} 
& 23.645 \textcolor{green}{(+0.61)} 
& 44.270 \textcolor{green}{(+0.66)} 
& 35.166 \textcolor{gray}{(-0.80)} 
& 20.714 \textcolor{gray}{(-3.52)} \\
\cline{2-10}
& \multirow{2}{*}{DanceTrack}
& \xcircle & DFINE & 34.269 & 29.645 & 84.107 & 70.189 & 16.950 & 25.04\\   [ -0.6ex]
& & \gcircle & DepTR 
& 36.712 \textcolor{green}{(+2.44)}
& 37.796 \textcolor{green}{(+8.15)}
& 83.267 \textcolor{gray}{(-0.84)}
& 67.588 \textcolor{gray}{(-2.60)}
& 20.121 \textcolor{green}{(+3.17)}
& 21.47 \textcolor{gray}{(-3.57)}\\
\Xhline{1.05pt}

\rowcolor[HTML]{FFFBE6}
\multicolumn{4}{c|}{\rule{0pt}{2.ex} \textbf{w/ DepTR-MOT (Avg. Gain $\Delta$)}} & 
\textbf{2.41} {\scriptsize (6.63\% $\uparrow$} & 
\textbf{3.59} {\scriptsize (9.75\% $\uparrow$} & 
\textbf{0.69} {\scriptsize (0.69\% $\uparrow$} & 
\textbf{0.36} {\scriptsize (0.06\% $\uparrow$} & 
\textbf{3.66} {\scriptsize (13.83\% $\uparrow$} & 
\textbf{5.21} {\scriptsize (17.09\% $\downarrow$} \\
\Xhline{1.05pt}
\cline{1-10}

\end{tabular}
\vskip-1ex
\caption{Results on QuadTrack, DanceTrack, SportsMOT, and VisDrone19. DepTR-MOT delivers robust performance improvements across diverse scenarios for major TBD trackers, with particularly substantial gains in IDF1 and AssA.}
\label{tab:results}
\vskip-5ex
\label{tab:main_table}

\end{table*}
Given depth maps $\mathbf{D} = \{ \mathbf{D}_{b,t} \}_{b=1}^{B}{}_{t=1}^{T}$, where each $\mathbf{D}_{b,t} \in \mathbb{R}^{1 \times H \times W}$ denotes the depth map of frame $t$ in batch $b$.
For each instance with a given 2D annotation box $B_i$, we employ a segmentation function $\mathcal{S}_{\theta}$ to generate a pixel-level mask:
\begin{equation}
M_i = \mathcal{S_{\theta}}(D_{b,t}, B_i), \quad M_i \in \{0,1\}^{H \times W}. \label{TFEq2}
\end{equation}
Using the element-wise multiplication $\odot$, the corresponding instance region depth $D_i$ is then obtained as
\begin{equation}
D_i = D_{b,t} \odot M_i,
\end{equation}
and the target-specific depth soft-label $Y_i$ is obtained by masked averaging,
\begin{equation}
Y_i = \frac{\sum_{(x,y)} D_{b,t}(x,y) \cdot M_i(x,y)}{\sum_{(x,y)} M_i(x,y)}.
\end{equation}
$Y_i$ denotes the instance-level depth supervision for each instance $i$, without interference from surrounding objects.

\noindent\textbf{{(3) Depth feature distillation guidance.}} To embed instance-level depth awareness into a DETR-style detector encoder, outputs are projected into $\mathbf{F_{src}} \in \mathbb{R}^{(B\cdot T)\times N \times C}$ via a lightweight MLP and aligned with teacher-generated features $\mathbf{F_{tea}}$. The alignment loss is defined as the multi-scale cosine similarity, averaged across scales $S$, frames $T$, and batches $B$, where $\langle \cdot,\cdot \rangle_F$ denotes the Frobenius inner product.

\begin{equation}
\begin{aligned}
\varepsilon_{b,t,s} =
1 - 
\frac{
    \left\langle F_{\mathrm{src},b,t,s}, 
    F_{\mathrm{tea},b,t,s} \right\rangle_F
}{
    \left\|F_{\mathrm{src},b,t,s}\right\|_F
    \cdot
    \left\|F_{\mathrm{tea},b,t,s}\right\|_F
},
\end{aligned}
\end{equation}
\begin{equation}
\mathcal{L}_{\mathrm{align}} = 
\frac{1}{BTS}
\sum_{b=1}^{B}
\sum_{t=1}^{T}
\sum_{s=1}^{S}
\varepsilon_{b,t,s}.
\label{eq:align_loss}
\end{equation}

To further guarantee numerical precision of the depth estimation, the predicted per-instance depth $P_i$ is directly compared against the teacher-generated depth $Y_i$ using a mean squared error to compute Depth Regression Loss:
\begin{equation}
\mathcal{L}_{\text{reg}} = \frac{1}{N}\sum_{i=1}^{N} \| P_i - Y_i \|_2^2,
\label{eq:depth_loss}
\end{equation}
where $N$ is the number of all instances. 
By jointly distilling these two losses, DepTR attains robust depth feature representations at the distributional level and depth predictions at the numerical level, thereby supporting our efficient framework, where $P$ denotes the predicted depth value for each instance.

\subsection{Mechanisms of Depth Decoding}
\label{subsec:Mechanisms_of_Depth_Decoding}
To endow the decoder with depth-sensing capabilities while maintaining structural simplicity, we extend the DETR branch with plug-in depth-aware layers, formulating depth decoding as an iterative, reference-guided refinement process. In this process, the high-precision positional outputs from the object localization decoder are utilized as candidate references to guide depth decoding. All modifications are built upon the standard DETR decoding layer~\cite{peng2024dfine} with minimal adaptation and lightweight integration.

Specifically, for the $q$-th query in the $b$-th batch, the predicted box center $c_{b,q}$ is transformed into a normalized reference anchor $r_{b,q}=\mathcal{T}(c_{b,q})$ to guide depth-aware cross-attention. At decoder layer $\ell$, the depth-aware attention $\mathcal{A}_d(\cdot)$ aggregates object-centric features using the previous query embedding and depth estimate:
\begin{equation}
\tilde{O}^{\ell}_{b,q} =
\mathcal{A}_d
\left(
O^{\ell-1}_{b,q}, V_b, r_{b,q}, p^{\ell-1}_{b,q}
\right),
\end{equation}
where $V_b$ denotes the multi-scale memory features and $p^{\ell-1}_{b,q}$ is the prior depth prediction.

The query representation and depth prediction are then iteratively refined as:
\begin{equation}
O^{\ell}_{b,q} =
\mathcal{T}_r
\left(
\tilde{O}^{\ell}_{b,q}, O^{\ell-1}_{b,q}
\right),
\end{equation}
\begin{equation}
p^{\ell}_{b,q}
=
p^{\ell-1}_{b,q}
+
f_d
\left(
O^{\ell}_{b,q}
\right),
\quad
p^{0}_{b,q}=0.
\end{equation}
After the final decoder layer, the depth prediction \(p_{b,q}\) matched by box-based Hungarian assignment, denoted as \(P_i\), is supervised by the corresponding teacher-generated depth label \(Y_i\) according to Eq.~\eqref{eq:depth_loss}.

The overall training objective is formulated as a combination of the alignment loss $\mathcal{L}_\text{align}$ (Eq.~\eqref{eq:align_loss}) and the standard position loss $\mathcal{L}_\text{box}$~\cite{zhu2020deformable}:
\begin{equation}
\mathcal{L} = \lambda_\text{box} \, \mathcal{L}_\text{box} 
            + \lambda_\text{reg} \, \mathcal{L}_\text{reg} 
            + \lambda_\text{align} \, \mathcal{L}_\text{align},
\end{equation}
where the coefficients \(\lambda_\text{box}, \lambda_\text{reg}, \lambda_\text{align}\) control the relative importance of box, depth, and alignment losses, respectively.

\begin{figure}[!t]
    \raggedleft  
    \includegraphics[width=0.49\textwidth]{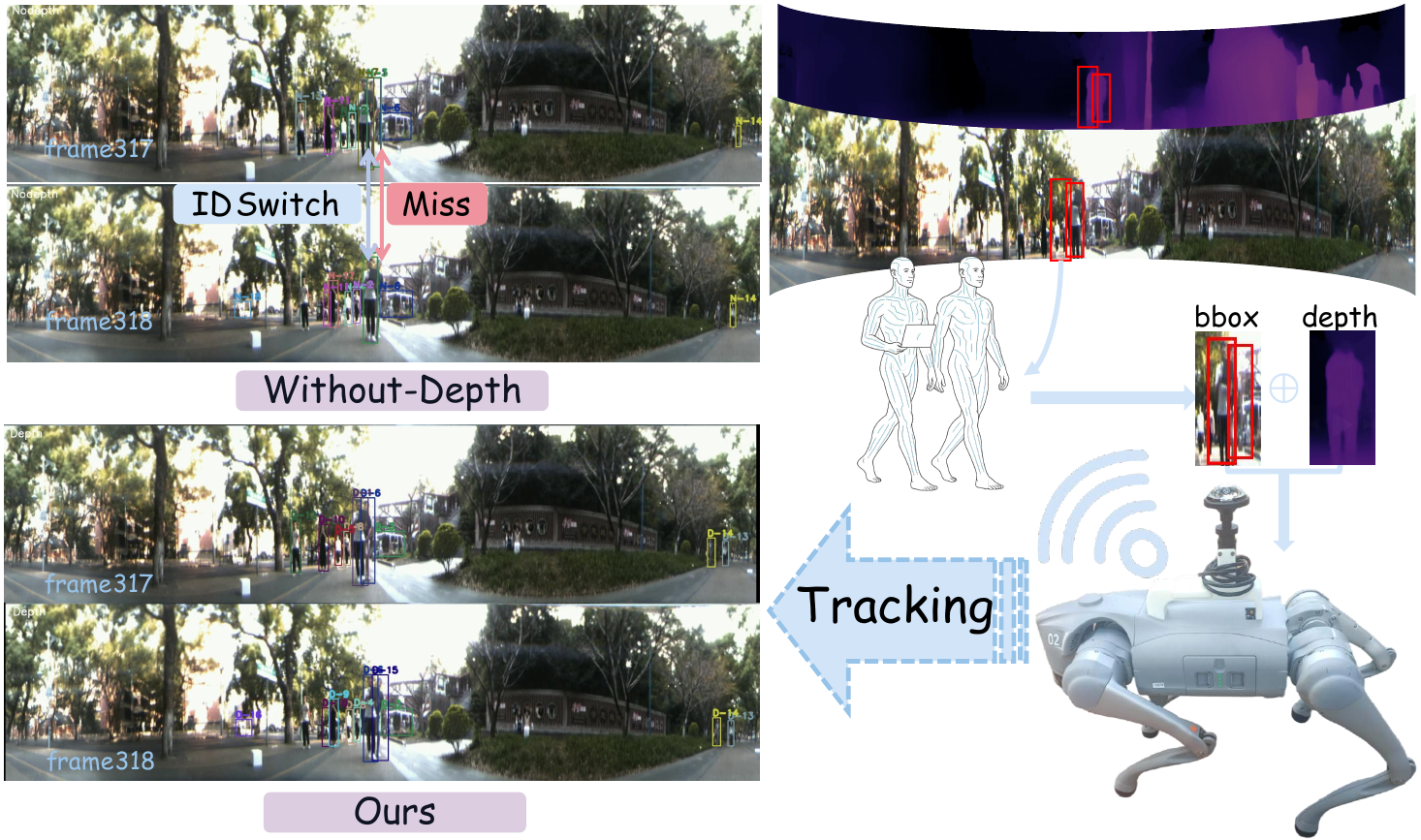} 
    \vskip-1ex
    \caption{Real-world application of DepTR-MOT: Deployed on a quadruped robot equipped with a panoramic camera, evaluating pedestrian tracking performance in sidewalk scenarios.}
    \label{fig:quadtrack.pdf}
    \vskip-4ex
\end{figure}

\subsection{Depth-Aware Association.}
\label{subsec:Depth_Aware_Association}
To enhance trajectory association, we design a depth-aware distance metric that computes normalized depth discrepancies for each pair.  
This depth distance cost matrix is fused into the second-stage matching pipeline, enabling the tracker to leverage complementary 3D geometry and appearance cues.
Such integration improves association reliability in a computationally efficient manner, especially in challenging scenarios with occlusions and densely crowded scenes.

Given a set of tracks $\mathcal{T}=\{t_i\}_{i=1}^M$ and detections $\mathcal{D}=\{d_j\}_{j=1}^N$ with depth estimates $P_{t_i}, P_{d_j}\in\mathbb{R}$, where $P_{t_i}$ is the depth prediction from the most recent valid frame of track $t_i$, we define the depth distance matrix as
\begin{equation}
D_{ij} =  \eta  {|\, P_{t_i} - P_{d_j} \,|}, 
\quad D \in [0,1]^{M \times N},
\end{equation}
where $\eta$ is the scale factor, acting as a normalization constant to normalize the values into the range $[0,1]$, which is set to 250 as the default setting according to~\cite{video_depth_anything}.

This matrix is integrated into the second-stage original matching cost:
\begin{equation}
C_{ij}' = \lambda  C_{ij} + \gamma D_{ij},
\label{eq:Fusion_Hyperparameters}
\end{equation}
where $\lambda$ and $\gamma$ are weighting factors, allowing the matching matrix to be adaptively refined.

\section{Experiments}
\subsection{Experiment Setup}
\subsubsection{Datasets}
To comprehensively evaluate our approach, we conduct experiments on four diverse multi-object tracking datasets: DanceTrack~\cite{sun2022dancetrack}, SportsMOT~\cite{cui2023sportsmot}, VisDrone2019~\cite{zhu2021detection}, and QuadTrack~\cite{luo2025omnidirectional}. Although some benchmarks are person-dominant, DepTR-MOT is category-agnostic and is applied to all annotated target categories defined by each dataset, without relying on human-specific appearance, pose, or motion priors. In addition to public benchmarks, we build a quadruped-robot panoramic perception system for real-world testing. The system is equipped with a panoramic camera and is used to collect sidewalk sequences with close pedestrian interactions and partial occlusions. These real-world sequences are not used for training, but only for deployment-oriented testing and qualitative analysis.

Both DanceTrack (100 sequences) and SportsMOT feature highly non-linear motion, severe occlusions, and indistinguishable target appearances (e.g., uniform clothing), making spatial-temporal association crucial. VisDrone2019 introduces UAV-captured aerial views, challenging the tracker with dense, small-scale targets and drastic perspective shifts.

In contrast, QuadTrack serves as our primary benchmark tailored for real-world robotic applications. It contains 32 panoramic sequences (17 train, 15 test) captured at 10 FPS on quadruped robots. Characterized by rapid target motion, complex outdoor environments, and severe depth ambiguity under panoramic distortion, QuadTrack perfectly encapsulates the challenges of close-proximity interactions and dynamic viewpoints. Compared to the others, it rigorously validates the robustness and effectiveness of our depth-aware framework in practical deployment scenarios.

\subsubsection{Metrics}
To evaluate final tracking performance, we adopt widely used multi-object tracking metrics, including Higher Order Tracking Accuracy (HOTA)\cite{luiten2021hota}, ID-based F1 Score (IDF1)\cite{ristani2016performance}, Association Accuracy (AssA)~\cite{bernardin2008evaluating}, and Multi-Object Tracking Accuracy (MOTA). HOTA provides a comprehensive assessment of detection and association quality, while IDF1 and AssA emphasize association consistency. In contrast, MOTA focuses mainly on detection accuracy.

\subsubsection{Implementation Details}
We use HGNetv2~\cite{chen2024hgnet} as the backbone, with the lightweight variant of Video Depth Anything~\cite{yang2024depthanyvideo} and SAM2~\cite{ravi2024sam2}, which already provides stable target-level absolute depth signals for supervision; increasing the teacher scale brings no consistent tracking gain, as DepTR-MOT mainly exploits region-averaged depth as an association-discriminative cue rather than for high-fidelity metric depth reconstruction. The input resolutions are $1078 \times 1918$ for DanceTrack~\cite{sun2022dancetrack}, $480 \times 2048$ for QuadTrack~\cite{luo2025omnidirectional}, $720 \times 1280$ for SportsMOT~\cite{cui2023sportsmot}, and $756 \times 1344$ for VisDrone19~\cite{zhu2021detection}. All models are trained for $5$ epochs without extra data augmentation, using AdamW with an initial learning rate of $2.5 \times 10^{-4}$ and batch size $1$. Experiments are conducted on two NVIDIA GeForce RTX 3090 GPUs with PyTorch 2.7.1 and CUDA 12.6.

\begin{table}[t!]
    \centering
    \setlength{\tabcolsep}{6pt} 
    \renewcommand{\arraystretch}{1.2}%
    \setlength{\extrarowheight}{-1pt}
    \scalebox{1.05}{
    \begin{tabular}{cc|ccccc}
        \Xhline{1.05pt}
        \rowcolor{mygray}
        $\lambda$ & $\gamma$ & HOTA $\uparrow$ & IDF1 $\uparrow$ & MOTA $\uparrow$ & DetA $\uparrow$ & AssA $\uparrow$ \\
        \Xhline{1.05pt}
1.0 & 0 & 23.666 & 22.201 & -2.6575 & 30.044 & 19.342 \\
0.1 & 0.9 & 13.384 & 12.456 & -65.001 & 17.523 & 11.555 \\
0.2 & 0.8 & 17.492 & 17.012 & -47.486 & 20.841 & 16.050 \\
0.3 & 0.7 & 20.623 & 20.655 & -36.736 & 22.973 & 20.095 \\
0.4 & 0.6 & 20.747 & 20.663 & -32.180 & 23.896 & 19.461 \\
0.5 & 0.5 & 22.553 & 22.835 & -29.161 & 24.630 & 22.514 \\
0.6 & 0.4 & 24.782 & 26.071 & -23.292 & 25.663 & 25.738 \\
0.7 & 0.3 & 26.299 & \textbf{28.769} & -7.920 & 29.048 & 24.987 \\

0.9 & 0.1 & 24.743 & 23.745 & \textbf{-2.109} & \textbf{30.257} & 21.031 \\
\rowcolor{mygray}0.8 & 0.2 & \textbf{27.590} & 28.035 & -3.655 & 29.957 & \textbf{26.553} \\
    \Xhline{1.05pt}
    \end{tabular}
    }
    \caption{Performance of DepTR-MOT on QuaTrack under different $\lambda$ and $\gamma$ weight configurations with ByteTrack~\cite{zhang2022bytetrack}.}
    \vspace{-2mm}
    \label{tab:ablation_tracker}
\vskip-3ex
\end{table}

\subsection{Quantitative Comparison}
Representative TBD trackers~\cite{wojke2017simple,zhang2022bytetrack,di2025hybridtrack,cao2023observation,Bewley2016_sort}, as summarized in Tab.~\ref{tab:main_table}. 
On these datasets, DepTR-MOT yields an average improvement of $+2.14$ HOTA, $+3.15$ IDF1, $+1.516$ MOTA, $+3.167$ AssA across different trackers, with only 6.57M depth-related parameters (total 38.45 M) introduced.
We observe significant improvements in IDF1, HOTA, and AssA, while DetA and MOTA change only marginally. Since DetA reflects detection quality, its marginal change indicates detection is largely unchanged, while gains in association-sensitive metrics (IDF1, AssA, HOTA) demonstrate improved trajectory continuity. The consistent gains in these association-sensitive metrics demonstrate that incorporating depth cues effectively enhances trajectory continuity. The only slight degradation appears for OC-SORT on DanceTrack, where MOTA and DetA improve but HOTA, IDF1, and AssA decrease marginally. This suggests that the fixed depth cost may occasionally interfere with OC-SORT’s motion association, and adaptive depth-confidence weighting is a promising future direction.

The limited improvement in MOTA is expected, as it is dominated by detection errors (FP and FN) rather than association quality. Therefore, the results confirm that the performance gains stem from improved association consistency enabled by the depth cues introduced by DepTR, rather than detection refinement or random fluctuations, while satisfying the real-time requirement of $22$ FPS~\cite{wang2020towards}.

\subsection{Ablation Studies}

\subsubsection{DepTR Architecture}
The ablation results in Tab.~\ref{ablation_study} show that supervision with $L_{reg}$ and $L_{align}$ (Exp.~\textbf{(2)}) improves tracking over the baseline, demonstrating the effectiveness of these two designs. Introducing the depth-aware weight $W_d$ with preprocessed global labels (Exp.~\textbf{(3)}) brings a substantial gain, confirming the strong contribution of $W_d$. However, combining $W_d$ with only $L_{reg}$ or $L_{align}$ (Exp.~\textbf{(4)/(5)}) yields weaker performance. In contrast, the full model (Exp.~\textbf{(6)}) achieves the best results, indicating that feature and label alignment must be coupled for the model to learn depth-aware perception and thereby optimize association.

\subsubsection{Data Loading Strategy}
Depth estimation models are often sensitive to scale, which may cause noticeable variations in estimated object positions across adjacent frames. Multi-frame input can mitigate this issue by providing additional temporal context. To analyze its impact, we conduct an ablation on different data loading strategies, with results reported in Tab.~\ref{tab:ablation_window_length}. 
The results show that our model achieves comparable MOT performance across all settings, indicating that DepTR is insensitive to input loading strategies.

\subsubsection{Fusion Hyperparameters}
As defined in Eq.~\ref{eq:Fusion_Hyperparameters}, depth cues are integrated into the matching cost matrix through two hyperparameters, $\lambda$ and $\gamma$, which balance position and depth. 
To evaluate their impact, we perform an ablation on the QuadTrack dataset using ByteTrack, with results shown in Tab.~\ref{tab:ablation_tracker}. The results indicate that when $\gamma < 0.4$, incorporating depth cues consistently improves tracking over the baseline, demonstrating that DepTR provides effective depth guidance under a broad range of settings. In contrast, when $\gamma > 0.4$, performance drops below the baseline, as excessive weighting on depth cues overemphasizes targets with similar depth while neglecting spatial relations, leading to incorrect associations and degraded tracking accuracy. For fair comparison, all compared experiments in this paper adopt the same $\lambda$ and $\gamma$, varying only the difference ablated in each experiment.

\subsection{Qualitative Results in Real-World Deployment}

To examine its practical value, we build a quadruped-robot panoramic perception system and deploy DepTR-MOT on real-world sidewalk sequences collected by this platform. The test scenes contain close-proximity pedestrian interactions, partial occlusions, and robot-viewpoint changes, which differ from controlled benchmark settings. As shown in Fig.~\ref{fig:quadtrack.pdf}, DepTR-MOT reduces trajectory fragmentation and identity switches by leveraging depth cues, maintaining consistent identities under challenging real-world conditions. These results demonstrate the robustness of DepTR-MOT for robotic perception tasks.

\begin{table}[t!]
\small
\renewcommand{\arraystretch}{1.2}
\setlength{\tabcolsep}{4pt}
\centering
\scalebox{0.85}{%
\begin{tabular}{c |ccc | ccccc}
\Xhline{1.05pt}
\rowcolor{mygray}  Exp. & $W_d$ & $L_{reg}$ & $L_{align}$ & HOTA $\uparrow$ & IDF1$\uparrow$ & MOTA $\uparrow$ & DetA$\uparrow$ & AssA $\uparrow$ \\
\Xhline{1.05pt}
(1) & -- & -- & --  & 42.343  & 83.352 & 26.640 & 67.691 & 45.922 \\
(2) & -- & \checkmark & \checkmark & 43.336 & 82.204 & 27.955 & 67.617 & 44.907 \\
(3) & \checkmark & -- & --  & 44.669 & 86.126 & 28.859 & 69.515 & 47.176 \\
(4) & \checkmark & -- & \checkmark & 43.722 & 85.705 & 27.779 & 69.201 & 46.231 \\
(5) & \checkmark & \checkmark & -- & 43.086 & 85.804 & 26.875 & 69.464 & 44.770 \\
\rowcolor{mygray}  
(6) & \checkmark & \checkmark & \checkmark & \textbf{44.939} & \textbf{86.271} & \textbf{29.133} & \textbf{69.687} & \textbf{47.956} \\
\Xhline{1.05pt}
\end{tabular}}
\caption{Ablation study on the key design components of DepTR on the DanceTrack dataset.}
\label{ablation_study}
\vskip-4ex
\end{table}

\section{Conclusion}
In this work, we present DepTR-MOT, a depth-aware detector tailored for the TBD paradigm that predicts instance-level depth at detection time and integrates seamlessly with existing trackers. Unlike conventional 2D detectors, DepTR adopts soft-label supervision from foundation models and dense depth distillation to learn reliable depth cues without requiring 3D annotations, while preserving the inference complexity of standard 2D detectors.

Experiments across diverse platforms, including quadruped robots, drones, and handheld or static cameras, show that DepTR-MOT improves association robustness and trajectory continuity under occlusions and close interactions while supporting real-time deployment. Future work will extend our training strategy from TBD-based trackers to end-to-end tracking frameworks, further exploring depth-informed MOT.

\bibliographystyle{IEEEtran}
\bibliography{references}

\begin{thebibliography}{10}
\providecommand{\url}[1]{#1}
\csname url@samestyle\endcsname
\providecommand{\newblock}{\relax}
\providecommand{\bibinfo}[2]{#2}
\providecommand{\BIBentrySTDinterwordspacing}{\spaceskip=0pt\relax}
\providecommand{\BIBentryALTinterwordstretchfactor}{4}
\providecommand{\BIBentryALTinterwordspacing}{\spaceskip=\fontdimen2\font plus
\BIBentryALTinterwordstretchfactor\fontdimen3\font minus \fontdimen4\font\relax}
\providecommand{\BIBforeignlanguage}[2]{{%
\expandafter\ifx\csname l@#1\endcsname\relax
\typeout{** WARNING: IEEEtran.bst: No hyphenation pattern has been}%
\typeout{** loaded for the language `#1'. Using the pattern for}%
\typeout{** the default language instead.}%
\else
\language=\csname l@#1\endcsname
\fi
#2}}
\providecommand{\BIBdecl}{\relax}
\BIBdecl

\bibitem{hu2023_uniad}
Y.~Hu \emph{et~al.}, ``Planning-oriented autonomous driving,'' in \emph{Proc. CVPR}, 2023, pp. 17\,853--17\,862.

\bibitem{li2025review}
S.~Li, H.~Ren, X.~Xie, and Y.~Cao, ``A review of multi-object tracking in recent times,'' \emph{IET Computer Vision}, vol.~19, no.~1, p. e70010, 2025.

\bibitem{milan2016mot16}
A.~Milan, L.~Leal-Taix{\'e}, I.~Reid, S.~Roth, and K.~Schindler, ``{MOT16:} {A} benchmark for multi-object tracking,'' \emph{arXiv preprint arXiv:1603.00831}, 2016.

\bibitem{sun2022dancetrack}
P.~Sun \emph{et~al.}, ``{DanceTrack:} {Multi-object} tracking in uniform appearance and diverse motion,'' in \emph{Proc. CVPR}, 2022, pp. 20\,961--20\,970.

\bibitem{yang2023hard}
F.~Yang, S.~Odashima, S.~Masui, and S.~Jiang, ``Hard to track objects with irregular motions and similar appearances? {Make} it easier by buffering the matching space,'' in \emph{Proc. WACV}, 2023, pp. 4788--4797.

\bibitem{zhao2025detrack}
W.~Zhao, Y.~Jiang, Y.~Gao, J.~Li, and X.~Gao, ``{DETrack:} {Depth} information is predictable for tracking,'' \emph{Neurocomputing}, vol. 616, p. 128906, 2025.

\bibitem{di2025hybridtrack}
L.~Di~Bella, Y.~Lyu, B.~Cornelis, and A.~Munteanu, ``{HybridTrack:} {A} hybrid approach for robust multi-object tracking,'' \emph{IEEE Robotics and Automation Letters}, vol.~10, no.~7, pp. 7238--7245, 2025.

\bibitem{zhou2018voxelnet}
Y.~Zhou and O.~Tuzel, ``{VoxelNet:} {End-to-end} learning for point cloud based {3D} object detection,'' in \emph{Proc. CVPR}, 2018, pp. 4490--4499.

\bibitem{weng2020ab3dmot}
X.~Weng, J.~Wang, D.~Held, and K.~Kitani, ``{AB3DMOT:} {A} baseline for {3D} multi-object tracking and new evaluation metrics,'' \emph{arXiv preprint arXiv:2008.08063}, 2020.

\bibitem{zuo2024towards}
Y.~Zuo, K.~Kayan, M.~Wang, K.~Jeon, J.~Deng, and T.~L. Griffiths, ``Towards foundation models for {3D} vision: How close are we?'' in \emph{Proc. 3DV}, 2025, pp. 1285--1296.

\bibitem{brazil2023omni3d}
G.~Brazil, A.~Kumar, J.~Straub, N.~Ravi, J.~Johnson, and G.~Gkioxari, ``{Omni3D:} {A} large benchmark and model for {3D} object detection in the wild,'' in \emph{Proc. CVPR}, 2023, pp. 13\,154--13\,164.

\bibitem{liu2023zero1to3}
R.~Liu, R.~Wu, B.~Van~Hoorick, P.~Tokmakov, S.~Zakharov, and C.~Vondrick, ``Zero-1-to-3: Zero-shot one image to {3D} object,'' in \emph{Proc. ICCV}, 2023, pp. 9264--9275.

\bibitem{10851814}
X.~Cao, Y.~Zheng, Y.~Yao, H.~Qin, X.~Cao, and S.~Guo, ``{TOPIC:} {A} parallel association paradigm for multi-object tracking under complex motions and diverse scenes,'' \emph{IEEE Transactions on Image Processing}, vol.~34, pp. 743--758, 2025.

\bibitem{cui2023sportsmot}
Y.~Cui, C.~Zeng, X.~Zhao, Y.~Yang, G.~Wu, and L.~Wang, ``{SportsMOT:} {A} large multi-object tracking dataset in multiple sports scenes,'' in \emph{Proc. ICCV}, 2023, pp. 9887--9897.

\bibitem{wang2025pd}
Y.~Wang, D.~Zhang, R.~Li, Z.~Zheng, and M.~Li, ``{PD-SORT:} {Occlusion-robust} multi-object tracking using pseudo-depth cues,'' \emph{IEEE Transactions on Consumer Electronics}, vol.~71, no.~1, pp. 165--177, 2025.

\bibitem{limanta2024camot}
F.~Limanta, K.~Uto, and K.~Shinoda, ``{CAMOT:} {Camera} angle-aware multi-object tracking,'' in \emph{Proc. WACV}, 2024, pp. 6465--6474.

\bibitem{quach2024depth}
K.~G. Quach, P.~Nguyen, C.~N. Duong, T.~D. Bui, and K.~Luu, ``Depth perspective-aware multiple object tracking,'' in \emph{Engineering Applications of AI and Swarm Intelligence}, 2024, pp. 181--205.

\bibitem{sun2025view}
H.~Sun, Y.~Li, G.~Yang, Z.~Su, and K.~Luo, ``View adaptive multi-object tracking method based on depth relationship cues,'' \emph{Complex \& Intelligent Systems}, vol.~11, no.~2, p. 145, 2025.

\bibitem{wu2024depthmot}
J.~Wu and Y.~Liu, ``{DepthMOT:} {Depth} cues lead to a strong multi-object tracker,'' \emph{arXiv preprint arXiv:2404.05518}, 2024.

\bibitem{zhao2024detrs}
Y.~Zhao \emph{et~al.}, ``{DETRs} beat {YOLOs} on real-time object detection,'' in \emph{Proc. CVPR}, 2024, pp. 16\,965--16\,974.

\bibitem{depth_anything_v2}
L.~Yang \emph{et~al.}, ``Depth anything {V2},'' in \emph{Proc. NeurIPS}, 2024, pp. 21\,875--21\,911.

\bibitem{ravi2024sam2}
N.~Ravi \emph{et~al.}, ``{SAM} 2: {Segment} anything in images and videos,'' in \emph{Proc. ICLR}, 2025.

\bibitem{peng2024dfine}
Y.~Peng, H.~Li, P.~Wu, Y.~Zhang, X.~Sun, and F.~Wu, ``{D-FINE:} {Redefine} regression task in {DETRs} as fine-grained distribution refinement,'' \emph{arXiv preprint arXiv:2410.13842}, 2024.

\bibitem{wojke2017simple}
N.~Wojke, A.~Bewley, and D.~Paulus, ``Simple online and realtime tracking with a deep association metric,'' in \emph{Proc. ICIP}, 2017, pp. 3645--3649.

\bibitem{luo2025omnidirectional}
K.~Luo \emph{et~al.}, ``Omnidirectional multi-object tracking,'' in \emph{Proc. CVPR}, 2025, pp. 21\,959--21\,969.

\bibitem{zhu2021detection}
P.~Zhu, L.~Wen, D.~Du, X.~Bian, H.~Fan, Q.~Hu, and H.~Ling, ``Detection and tracking meet drones challenge,'' \emph{IEEE Transactions on Pattern Analysis and Machine Intelligence}, vol.~44, no.~11, pp. 7380--7399, 2021.

\bibitem{girshick2014rich}
R.~B. Girshick, J.~Donahue, T.~Darrell, and J.~Malik, ``Rich feature hierarchies for accurate object detection and semantic segmentation,'' in \emph{Proc. CVPR}, 2014, pp. 580--587.

\bibitem{ren2016faster}
S.~Ren, K.~He, R.~Girshick, and J.~Sun, ``Faster {R-CNN:} {Towards} real-time object detection with region proposal networks,'' \emph{IEEE Transactions on Pattern Analysis and Machine Intelligence}, vol.~39, no.~6, pp. 1137--1149, 2017.

\bibitem{liu2025dual}
Z.~Liu, J.~Wu, Y.~Cai, H.~Wang, L.~Chen, and Q.~Liu, ``Dual-stage feature specialization network for robust visual object detection in autonomous vehicles,'' \emph{Scientific Reports}, vol.~15, no.~1, p. 15501, 2025.

\bibitem{liu2016ssd}
W.~Liu \emph{et~al.}, ``{SSD:} {Single} shot multibox detector,'' in \emph{Proc. ECCV}, 2016, pp. 21--37.

\bibitem{redmon2016you}
J.~Redmon, S.~Divvala, R.~Girshick, and A.~Farhadi, ``You only look once: Unified, real-time object detection,'' in \emph{Proc. CVPR}, 2016, pp. 779--788.

\bibitem{lin2025gridclip}
J.~Lin, S.~Sun, and S.~Gong, ``{GridCLIP:} {One-stage} object detection by grid-level {CLIP} representation learning,'' \emph{Pattern Recognition}, p. 112187, 2026.

\bibitem{carion2020end}
N.~Carion, F.~Massa, G.~Synnaeve, N.~Usunier, A.~Kirillov, and S.~Zagoruyko, ``End-to-end object detection with transformers,'' in \emph{Proc. ECCV}, 2020, pp. 213--229.

\bibitem{zhu2020deformable}
X.~Zhu, W.~Su, L.~Lu, B.~Li, X.~Wang, and J.~Dai, ``Deformable {DETR}: {Deformable} transformers for end-to-end object detection,'' in \emph{Proc. ICLR}, 2021.

\bibitem{song2021vidt}
H.~Song \emph{et~al.}, ``{ViDT:} {An} efficient and effective fully transformer-based object detector,'' in \emph{Proc. ICLR}, 2022.

\bibitem{cao2023observation}
J.~Cao, J.~Pang, X.~Weng, R.~Khirodkar, and K.~Kitani, ``Observation-centric {SORT}: {Rethinking} {SORT} for robust multi-object tracking,'' in \emph{Proc. CVPR}, 2023, pp. 9686--9696.

\bibitem{welch1995introduction}
G.~Bishop and G.~Welch, ``An introduction to the kalman filter,'' \emph{Proc of SIGGRAPH, Course}, vol.~8, no. 27599-23175, p.~41, 2001.

\bibitem{kuhn1955hungarian}
H.~W. Kuhn, ``The hungarian method for the assignment problem,'' \emph{Naval Research Logistics Quarterly}, vol.~2, no. 1-2, pp. 83--97, 1955.

\bibitem{lv2024diffmot}
W.~Lv, Y.~Huang, N.~Zhang, R.-S. Lin, M.~Han, and D.~Zeng, ``{DiffMOT:} {A} real-time diffusion-based multiple object tracker with non-linear prediction,'' in \emph{Proc. CVPR}, 2024, pp. 19\,321--19\,330.

\bibitem{zhang2022bytetrack}
Y.~Zhang \emph{et~al.}, ``{ByteTrack:} {Multi-object} tracking by associating every detection box,'' in \emph{Proc. ECCV}, 2022, pp. 1--21.

\bibitem{yang2024hybrid}
M.~Yang \emph{et~al.}, ``{Hybrid-SORT:} {Weak} cues matter for online multi-object tracking,'' in \emph{Proc. AAAI}, 2024, pp. 6504--6512.

\bibitem{khanchi2025depth}
M.~Khanchi, M.~Amer, and C.~Poullis, ``Depth-aware scoring and hierarchical alignment for multiple object tracking,'' in \emph{Proc. ICIP}, 2025.

\bibitem{video_depth_anything}
S.~Chen \emph{et~al.}, ``Video depth anything: Consistent depth estimation for super-long videos,'' in \emph{Proc. CVPR}, 2025, pp. 22\,831--22\,840.

\bibitem{Bewley2016_sort}
A.~Bewley, Z.~Ge, L.~Ott, F.~Ramos, and B.~Upcroft, ``Simple online and realtime tracking,'' in \emph{Proc. ICIP}, 2016, pp. 3464--3468.

\bibitem{luiten2021hota}
J.~Luiten \emph{et~al.}, ``{HOTA:} {A} higher order metric for evaluating multi-object tracking,'' \emph{International Journal of Computer Vision}, vol. 129, no.~2, pp. 548--578, 2021.

\bibitem{ristani2016performance}
E.~Ristani, F.~Solera, R.~Zou, R.~Cucchiara, and C.~Tomasi, ``Performance measures and a data set for multi-target, multi-camera tracking,'' in \emph{Proc. ECCVW}, 2016, pp. 17--35.

\bibitem{bernardin2008evaluating}
K.~Bernardin and R.~Stiefelhagen, ``Evaluating multiple object tracking performance: The {CLEAR} {MOT} metrics,'' \emph{EURASIP Journal on Image and Video Processing}, vol. 2008, no.~1, p. 246309, 2008.

\bibitem{chen2024hgnet}
Z.~Chen, C.~Tang, and L.~Xiong, ``{HGNET:} {A} hierarchical feature guided network for occupancy flow field prediction,'' \emph{arXiv preprint arXiv:2407.01097}, 2024.

\bibitem{yang2024depthanyvideo}
H.~Yang \emph{et~al.}, ``Depth any video with scalable synthetic data,'' \emph{arXiv preprint arXiv:2410.10815}, 2024.

\bibitem{wang2020towards}
Z.~Wang, L.~Zheng, Y.~Liu, Y.~Li, and S.~Wang, ``Towards real-time multi-object tracking,'' in \emph{Proc. ECCV}, 2020, pp. 107--122.

\end{thebibliography}

\end{document}